\documentclass[runningheads]{llncs}
\usepackage{graphicx}
\usepackage{subcaption}
\usepackage{wrapfig}
\usepackage{float}
\usepackage{commath}
\begin{document}
\title{PoshakNet: Framework for matching dresses from real-life photos using GAN and Siamese Network}
%
%
\author{Abhigyan Khaund\and
Daksh Thapar\and
Aditya Nigam}

\authorrunning{K. Abhigyan et al.}
\institute{Indian Institute of Technology, Mandi\\
\email{\{b16082,d18033\}@students.iitmandi.ac.in, aditya@iitmandi.ac.in}
}
\maketitle              
\begin{abstract}
Online garment shopping has gained many customers in recent years. Describing a dress using keywords does not always yield the proper results, which in turn leads to dissatisfaction of customers. A visual search based system will be enormously beneficent to the industry. Hence, we propose a framework that can retrieve similar clothes that can be found in an image. The first task is to extract the garment from the input image (street photo). There are various challenges for that, including pose, illumination, and background clutter. We use a Generative Adversarial Network for the task of retrieving the garment that the person in the image was wearing. It has been shown that GAN can retrieve the garment very efficiently despite the challenges of street photos. Finally, a siamese based matching system takes the retrieved cloth image and matches it with the clothes in the dataset, giving us the top $k$ matches. We take a pre-trained inception-ResNet v1 module as a siamese network (trained using triplet loss for face detection) and fine-tune it on the shopping dataset using center loss. The dataset has been collected inhouse. For training the GAN, we use the LookBook dataset~\cite{yoo2016pixel}, which is publically available.

\keywords{Deep Learning  \and GAN \and e-commerce. \and Siamese \and Dress Retrieval}
\end{abstract}
\section{Introduction}
Clothes are like an extended body part of human beings. Every human being spends hours every day to make themselves look good in the best attire. To achieve this, we spent a lot of time buying clothes that we fancy. Recent years have seen a tremendous rise in e-commerce ventures and their sales. Clothes shopping has led these sales after electronics. This wave of online apparel shopping is due to the involvement of tech giants like Amazon, Flipkart, Myntra, and others. Users of these platforms generally search for clothes using some static and manually selected keywords that describe the shape or color of the dress. However, it is not always that searching through such keywords leads users directly to the desired dress. 

\textbf{Motivation}: More often than not, we tend to search for clothes that we see others wearing, and we find it challenging to frame that dress in terms of the available keywords. It is difficult to describe multi-colored apparel, and with so many new fashion designs hitting the market, it is sometimes confusing to pinpoint the dress type worn. These issues result in users not being able to search for the desired dress in the shopping sites and often getting unwanted results. This is the major drawback of any keyword based matching algorithms and can be solved if the products can be described using visual clues. We base our search model not on keywords but directly on the image content. We take a picture of the dress the user wants to find matches for and run a matching process to determine the closest matches to the dress from the shopping portal, reducing the dependence on being able to search and shop with exact keywords and textual descriptions.

\begin{figure}
\centering
  \includegraphics[width=0.7\linewidth]{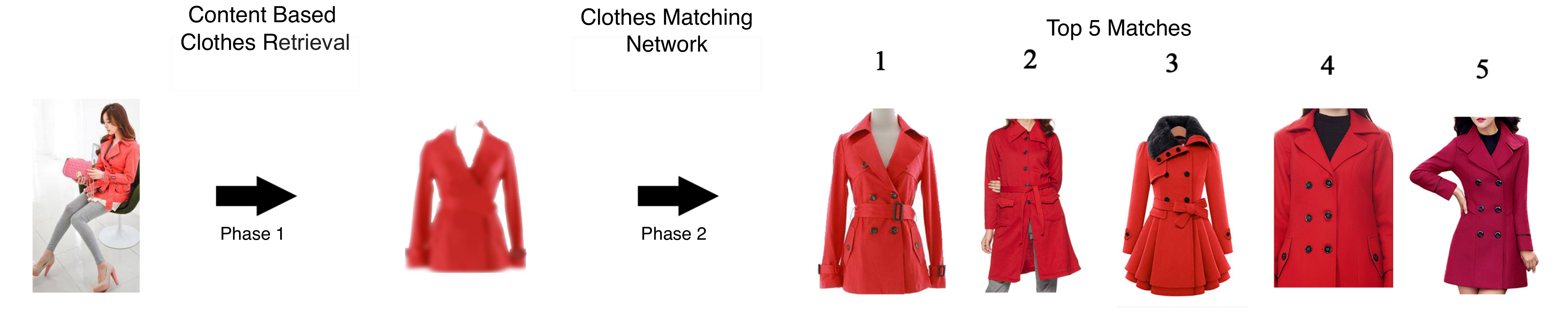}
  \caption{An example showing the end-to-end working of the street-to-shop problem.}
  \label{figure:problem}
\end{figure}
\begin{figure}
\vspace{-5mm}
\centering
  \includegraphics[width=\linewidth]{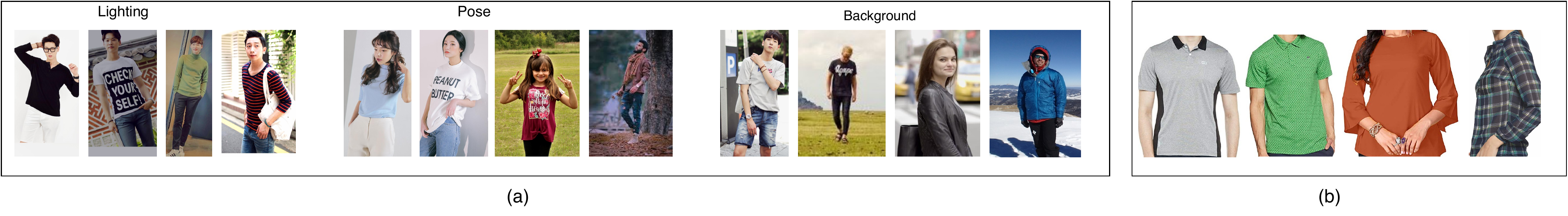}
  \caption{(a) Illustration of various challenges to content-based clothes retrieval. (b) Cropped picture from bounding box with unwanted body elements.}
  \label{figure:problem}
\end{figure}
\textbf{Challenges:} How do we get a visual representation of the dress in a way that we can accurately do its matching with the available shopping products? If we directly use a person's picture for matching, it is bound to fail, as such photos have a varying background, inconsistent lighting, and different positions of the person's posture, which may lead to an obscured view of the dress. One easy solution to this problem is to crop the image using object detection to retrieve only the bounding box for the dress of the person from the image~\cite{zhang2018watch}. This approach is unable to remove certain unwanted portions like arms, hand, neck, or hair in the cropped portion. If the person is also at an angle that causes the dress to be not fully visible, this approach only gets the visible part of the dress due to object detection. These issues cause significant performance degradation and may ruin a user's shopping experience and leave them completely unsatisfied. To handle this problem, we should be able to generate the image of the dress directly from the image without any unwanted background or noise. That is, when the user provides an image of a person, a new image is produced consisting only of the dress of the person.  The extraction should be independent of the posture of the person in the image, background, or any other noise.

\textbf{Problem Statement}: This work focuses on solving addressing above mentioned issues as a bi-phase problem and providing a single pipeline (shown in Fig.~\ref{figure:problem}) that does an end-to-end work for solving this issue of street-to-shop. The proposal consists of two phases:

\begin{enumerate}
    \item Content-Based Clothes Retrieval(CBCR) - Given an input street query photo with a single person in it, generate an image using a generative network with only the dress of the person worn in the input image.
    \item Clothes Matching Network(CMN) - Use the generated image to find a match in the shopping products dataset using a matching convolution network and show the users top k matches.
\end{enumerate}

\begin{table}
\vspace{-5mm}
    \begin{tabular}{|p{0.33\linewidth}|p{0.26\linewidth}|p{0.41\linewidth}|}
\hline
        Clothes Retrieval                                                                                                  & Product Similarity                                  & Street To Shop                                                                                                                                            \\ \hline
        Manual semantic attributes annotation, manual effort, dependent on correctness of semantic attributes~\cite{wan2014deep}. & Siamese Networks most popular to measure similarity of two images. & Retrieving clothes from daily photos and matching similar looking (or identical) products is called Street to Shop.                                        \\ 
        Deep learning models to handle cross-scenario variations, learn distinct features~\cite{gajic2018cross,huang2015cross,jiang2016deep}.                                  & Fine-grained object retrieval matching with triplet based ranking loss~\cite{lai2015simultaneous,wang2014learning}.      & Align body parts in the street photo with shop photo,  works in constrained environment, distorted results in real-life scenarios~\cite{liu2012street}. \\ 
        Object detection to get dress's bounding box~\cite{liang2016clothes,hadi2015buy}.                                                                       & Face matching using triplet loss to train~\cite{schroff2015facenet}.           & Use image annotation and a bounding box of the query product, lot of human interference required~\cite{hadi2015buy}.                                                          \\ 
        Generative Adversarial Nets(GAN)~\cite{goodfellow2014generative} generate clothes from person's picture~\cite{yoo2016pixel,zhang2018watch}.                                            & ~                                                   & Bounding box introduces unwanted human parts and background in the picture, contrast to clean shopping product image.                                      \\
\hline
\end{tabular}
\caption{Summary of the work in relevant literature.}
\vspace{-7mm}
\end{table}

\textbf{Contributions: }
This paper addresses the Street to Shop problem as a bi-phase problem using deep learning methods. Clothes retrieval is considered the first phase and is approached by attempting to generate a new image of the clothes from the input image. This generation is independent of the human posture or background of the input image. Then, we explore siamese networks to establish similarity between the generated and shopping image. This invloves learning similarity in the clothing domain for the network.

Specific contributions are:
\begin{itemize}
    \item Creation of a dataset, a shopping dataset that contains clean product images of the front view of the dress without any human part. Additionally, this dataset is classified based on fixed attributes manually. We will make the dataset publically available post-acceptance. 
    \item Transfer learning of a siamese network trained on face dataset to the clothing domain using center loss. To the best of our knowledge, this is the first attempt where a siamese network has been fine-tuned to generate top recommendations of clean product images without any human or mannequin body similar to a query image.   
    \item Exploring the utility of using a generative network for the clothes retrieval phase of the street to shop problem, constructing clean dress images invariant of the human pose, background, dress occlusion, and lighting in the input image.

\end{itemize}

The remaining paper is organized as: Section \ref{sec:method} describes the proposed methodology. Section \ref{sec:analysis} discuss the experimental analysis and Section \ref{sec:conc} presents the conclusion and scope of future work.
\section{Proposed Methodology}\label{sec:method}
\begin{figure}
\centering
\begin{subfigure}[t]{0.75\linewidth}
  \includegraphics[width=\linewidth]{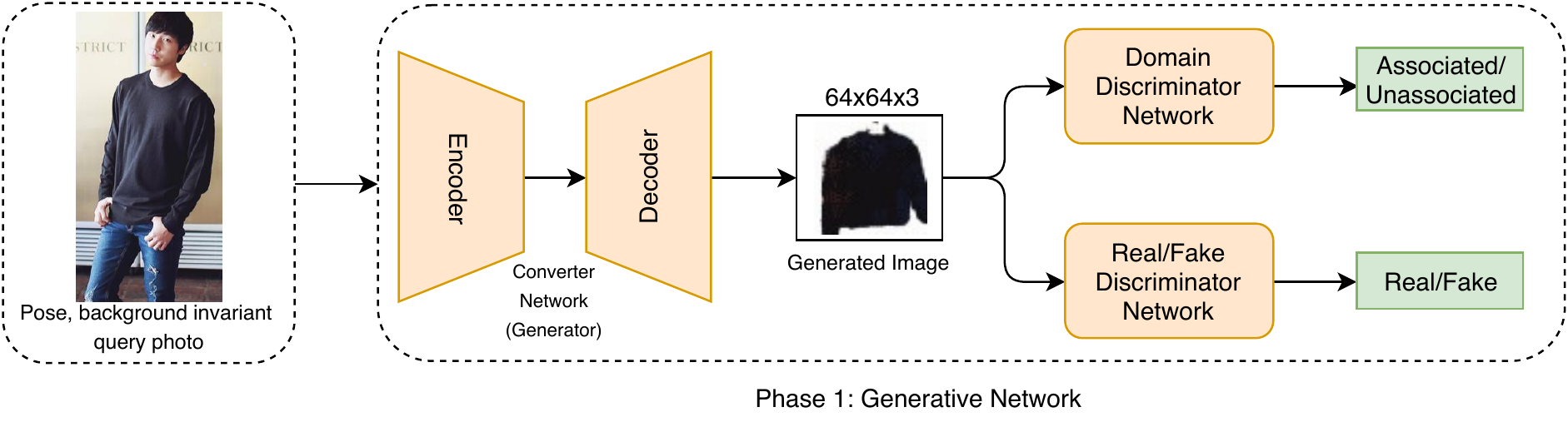}
  \caption{(a)}
  \label{figure:generator}
\end{subfigure}

\begin{subfigure}[t]{\textwidth}
\centering
  \includegraphics[width=0.40\linewidth]{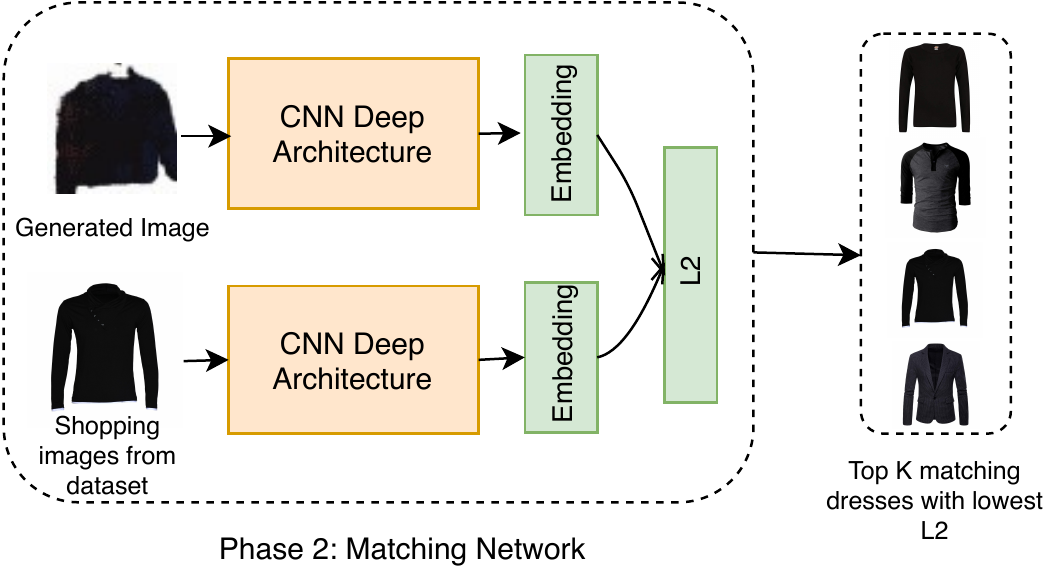}
  \caption{(b)}
  \label{figure:matching}
\end{subfigure}
\caption{(a) A Generative Network that is used generate the clean image of the product the person is wearing. (b) A matching network that matches the generated image with the images from the dataset.}
\end{figure}

This section covers the detailed description of the methodology proposed in this work. Our methodology is divided into 2 phases, clothes retrieval, and product matching.

\subsection{Phase 1: Content-Based Clothes Retrieval(CBCR)}
The first phase requires to generate an image of a dress from a street image containing a person wearing the dress. We propose a network that is invariant of the pose of the person and does not depend on the lighting conditions. It is also able to work if parts of the dress are occluded. The proposed idea is related to generative image models wherein we generate a final result as an image directly from an input image. There are two types of image-generative models, one with generative parametric approaches and one with adversarial approaches. In this work, we have used the adversarial approach. The adversarial approach was proposed by Goodfellow et al. as Generative Adversarial Nets (GAN)~\cite{goodfellow2014generative}. We want to use a GAN that works in the clothes domain and can generate clothes from the provided image. One approach to this is to do a pixel-level domain transfer in the GAN~\cite{yoo2016pixel} on the clothes domain. We make use of~\cite{yoo2016pixel} trained on the LookBook dataset. Using the generative network, we produce an image containing only the dress of size $64\times64\times3$. 
\begin{table}
\vspace{-4mm}
\begin{center}
    \begin{tabular}{|c|c|c|c|c|}
        \hline
        Layers         & Filters      & Number of Filters & Stride & Padding \\ \hline
        Convolution 1 & 5x5x\{3,3,6\} & 128               & 2      & 2   \\ 
        Convolution 2 & 5x5x128     & 256               & 2      & 2   \\ 
        Convolution 3 & 5x5x256     & 512               & 2      & 2   \\ 
        Output Layer  & 1x1x1024    & \{64, 1, 1\}        & 1      & 0   \\
        \hline
    \end{tabular}
    \caption{ Network layers of the \{encoder, real/fake discriminator, domain discriminator\} of Generative Network. \cite{yoo2016pixel}}
    \label{tab:generative_network}
\end{center}
\vspace{-12mm}

\end{table}

\begin{table}
\vspace{-10mm}

\begin{center}
    \begin{tabular}{|c|c|c|c|}
 \hline
        Layer              & Filter   & Number of Filters & Stride \\ \hline
        Convolution 1      & 1x1x64   & 4x4x1024          & 1      \\ 
        Conv2DTranspose. 1 & 5×5×512  & 1024              & 2      \\ 
        Conv2DTranspose. 2 & 5×5×256  & 512               & 2      \\ 
        Conv2DTranspose. 3 & 5×5×128  & 256               & 2      \\ 
        Conv2DTranspose. 4 & 5×5×3    & 128               & 2      \\
 \hline
\end{tabular}
    \caption{ Network layers of the decoder of Generative Network. \cite{yoo2016pixel}}
    \label{tab:generative_network_decoder}
\end{center}
\vspace{-9mm}
\end{table}

The architecture proposed involved in \cite{yoo2016pixel} involves a converter network and a discriminator network. The converter is a network consisting of two parts encoder and decoder. Both the encoder and decoder are composed of convolutional and transpose-convolutional layers, respectively. The encoder condenses the input to a 64-dimension capturing the semantic attributes, and the decoder then constructs the relevant target image from it. Table \ref{tab:generative_network} and Table \ref{tab:generative_network_decoder} describes the architectures of the encoder and decoder respectively. In the encoder network, the first four layers have L-ReLU as activation function, while the decoder network uses ReLU as the activation function in the first four layers.

On top of the converter network, it uses two discriminator networks, which behave as an adversary and guide the converter network. The first discriminator network $D_{R}$ is used to differentiate between fake and real images, fake being the one generated by the converter, and real are the actual dataset images. The second is a domain discriminator $D_{A}$, which produces a scalar probability specifying whether the input image and the generated image are associated or not. In case of an unrealistic generated image, the real/fake discriminator backpropagates a loss while in case of a generated image irrelevant to the input, the domain discriminator backpropagates a loss.  

The real/fake discriminator loss $L_{R}$ is a binary cross entropy loss, defined as - 
\begin{equation}\label{eqn:real}
    L_{R} = -t \cdot log[D_{R}(I)]+(t-1)\cdot log[1-D_{R}(I)]
\end{equation}
Here, $t$ is 1 if $I$ (input) is a real image from training set and 0 if $I$ is  a fake image drawn by the generator.
The domain discriminator loss $L_{A}$ is also defined similarly - 
\begin{equation}\label{eqn:domain}
    L_{A} = -t \cdot log[D_{A}(I_{S},I)]+(t-1)\cdot log[1-D_{A}(I_{S},I)]
\end{equation}
Here $I_{S}$ is the source image and $t$ is 1 if $I$ (input) is a ground truth target and 0 if $I$ is a irrelevant target or an inference from converter.

Considering both eqn. \ref{eqn:real} and \ref{eqn:domain}, the loss of the converter network $L_{C}$ is defined as - 
\begin{equation}\label{eqn:converter}
    L_{C}(I_{S},I) = -\frac{1}{2}L_{R}(I) - \frac{1}{2}L_{A}(I_{S},I),
\end{equation}
where $I$ is a random selection with equal probability among the ground truth, inference and irrelevant target.


\subsection{Phase 2: Clothes Matching Network(CMN)}
In this phase, we solve the matching problem using deep learning techniques. We use a pre-trained Inception ResNet v1 model \cite{szegedy2015going}, trained over face dataset using triplet loss function~\cite{schroff2015facenet} and do transfer learning on top of it.  The pre-trained model is capable of mapping an input image to a 128-d feature space, where each point lies on a unit radius hypersphere centered at the origin. Initially, it is trained to face datasets, but we have to retrain it in order to create a similarity measure for clothes. Hence, we fine-tuned the Inception ResNet v1 of Facenet using the shopping dataset (described in section~\ref{sec:daata}) we created. The fine-tuning is done on joint supervision of softmax loss and center loss function~\cite{wen2016discriminative}.  The center loss helps learn a center for the deep features of each class, moving the features of the same class to their centers. Softmax helps enlarge the inter-class difference  and the center loss help reduce the intra-class feature distance. 

The softmax loss $L_{S}$ function is defined as,
\begin{equation}\label{eqn:softmax}
    L_{S} = - {\sum}_{i=1}^{m}log{(e^{W^{T}_{y_{i}}x_{i}+b_{y_{i}}})\over {{\sum}_{j=1}^{n}e^{W^{T}_{j}x_{i}+b_{j}}}}
\end{equation}

In Eq. \ref{eqn:softmax}, $x_{i}$ denotes the $i^{th}$ deep feature from the $y$th class. $W_{j}$ is the value of $j^{th}$ column of the weight matrix $W$ of the the last layer and $b$ is the bias of that layer. $m$ is the size of mini-batch and $n$ denotes the number of classes.

To minimise intra-class variations of different classes, center loss~\cite{wen2016discriminative} $L_{C}$ is defined as,
\begin{equation}\label{eqn:center}
    L_{C} = \frac{1}{2} \sum_{i=1}^{m} \norm{x_{i} - c_{y_{i}}}^{2}_{2}
\end{equation}{}
Here, $c_{y_{i}}$ is the $y_{i}^{th}$ class center of the deep features. $x_{i}$ and $m$ are as defined for Eq. \ref{eqn:softmax}.

For joint supervision, the softmax loss(\ref{eqn:softmax}) and center loss(\ref{eqn:center}) are added to form the final loss to train the CNNs, $L$ = $L_{S}$ + $\lambda L_{C}$ where $\lambda$ is a scalar for balancing the two loss functions. The model is fine-tuned for 10 epochs with $\lambda$ value  of 0.95. 
For extracting the products that are similar to the image generated in phase1, we use a siamese based matching framework. The trained inception-ResNet v1 model is used to create a feature embedding of the generated image. This embedding is then matched with the embeddings of all the images that are present in the dataset by computing $L2$ distance score. The output score for each pair of the input image and the products are sorted, and the top $k$ results are shown as nearest to the input dress. 

\textbf{Justification:} We want those products which are of different categories have their deep vectors away from each other. Within a category, we want closer matching among similar dresses and yet be able to differentiate among different designs. Using joint supervision of softmax and center loss over triplet loss helps us achieve distinction between different products within the same category as we could treat each product as a class of its own. Intuitively, the softmax loss forces the deep features of the different products and categories apart, while the center loss pulls the same product images to a common center.

\section{Experimental Analysis}\label{sec:analysis}
For validating the performance of our proposed approach, we have used two datasets. LookBook dataset~\cite{yoo2016pixel} is used to train the image retrieval module, and the matching network is validated using our in house collected dataset. For validating the matching network, we have computed the precision in retrieving top $k$ similar images.
\subsection{Datasets Specification}
\label{sec:daata}
\begin{figure}
  \vspace{-6mm}
\centering
  \includegraphics[width=0.9\textwidth, height=0.6in]{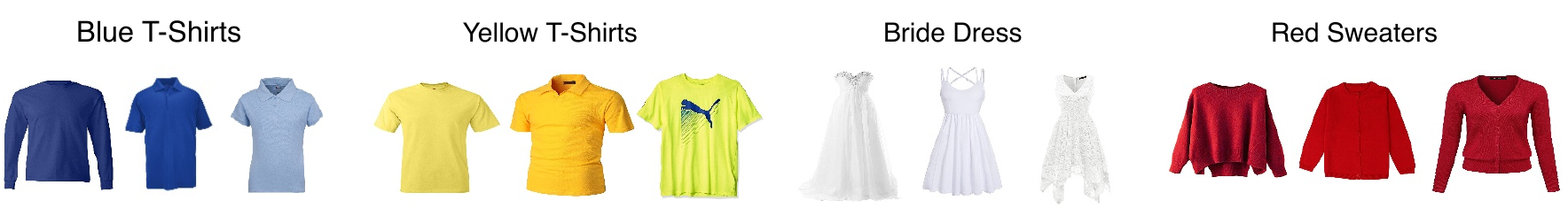}
    \caption{Collected shopping dataset examples.}
  \label{figure:dataset}
  \vspace{-8mm}
\end{figure}
\paragraph{GAN Dataset:}
For training the image retrieval system,  we have utilized the LookBook dataset~\cite{yoo2016pixel}. There are two categories of images in the dataset, one containing fashion models in different backgrounds and poses in a single dress, and the other is a corresponding product image with a clean background. The dataset consists of 84,748 images with 9,732 product images that are associated with 75,016 model images of the first category. Each model image on an average has eight pictures in different background and pose with the same product.  
\paragraph{Shopping Dataset:} We make a shopping dataset that contains images of products from the Amazon website. For every product, we take a product image with a clean background that contains the front view of the dress. We use data augmentation to generate eight other pictures of the same product. The products are of varied color and types. There are 326 products in the database. It means there are a total of  2,608 images. Five categories are used to classify the products. The categories are \textit{Blue T-Shirts}, \textit{Red Sweaters}, \textit{Bridal Dress}, \textit{Yellow T-Shirts} and \textit{Others}. \textit{Others} category consists of apparels that do not fall into any of the first four categories. We have used $80\%$ data from all the categories for training the matching network and the rest $20\%$ for testing.Fig.~\ref{figure:dataset} shows two examples from each of the categories in our dataset.

\subsection{Performance metrics}\label{sec:pm}

We follow ranking based evaluation criteria~\cite{liu2012street}. For an input image $i$, we are matching it to all the $n$ images in the \textit{shopping dataset}. The matches are ranked based on the score resulted from the matching network. The product with the lowest score is ranked the highest. We define a binary value $Res(r)$ which is the ground truth category relevance between $i$ and the $r\textsuperscript{th}$ ranked image. Specifically, if the ground truth category of the input image $i$ and the matched image are of the same category, then the $Res$ value is 1. If the categories differ, the $Res$ value takes 0.

We can evaluate a ranking of top $k$ retrieved product images with respect to an input $i$ by precision
\begin{equation}
    Precision@k ={{\sum}_{r}^{k}Rel(r)\over N}
\end{equation}
where $N$ is a normalization constant equal to $k$. This ensures that the correct ranking results in an precision score of 1.
\subsection{Results and analysis}
\begin{table}
\tiny
\centering
\begin{tabular}{|c|l|l|l|l|l|l|l|l|l|l|l|l|l|l|l|}
\hline
        $k$ & 1     & 2     & 3     & 4     & 5     & 6     & 7     & 8     & 9     & 10    & 11    & 12    & 13    & 14    & 15    \\ \hline
        $Precision@k$  & 0.846 & 0.846 & 0.820 & 0.811 & 0.793 & 0.764 & 0.742 & 0.713 & 0.687 & 0.667 & 0.646 & 0.625 & 0.608 & 0.596 & 0.576 \\
\hline
\end{tabular}
  \caption{Precision of the system with k varying from 1 to 15.}
  \label{tab:plot}
  \vspace{-8mm}
\end{table}
Performance is measured in terms of $Precision@k$ as described in Section~\ref{sec:pm}. Table.~\ref{tab:plot} depicts the result of a precision match with top $k$ products for $k$ values ranging from 1 to 15. Our approach yields a high precision of $0.84$ when $k$ is 1 and decreases gradually to $0.793$, $0.667$, and $0.576$ for $k$ values of 5, 10, and 15 respectively.

\begin{figure}[!ht]
\centering
  \includegraphics[width=0.5\linewidth]{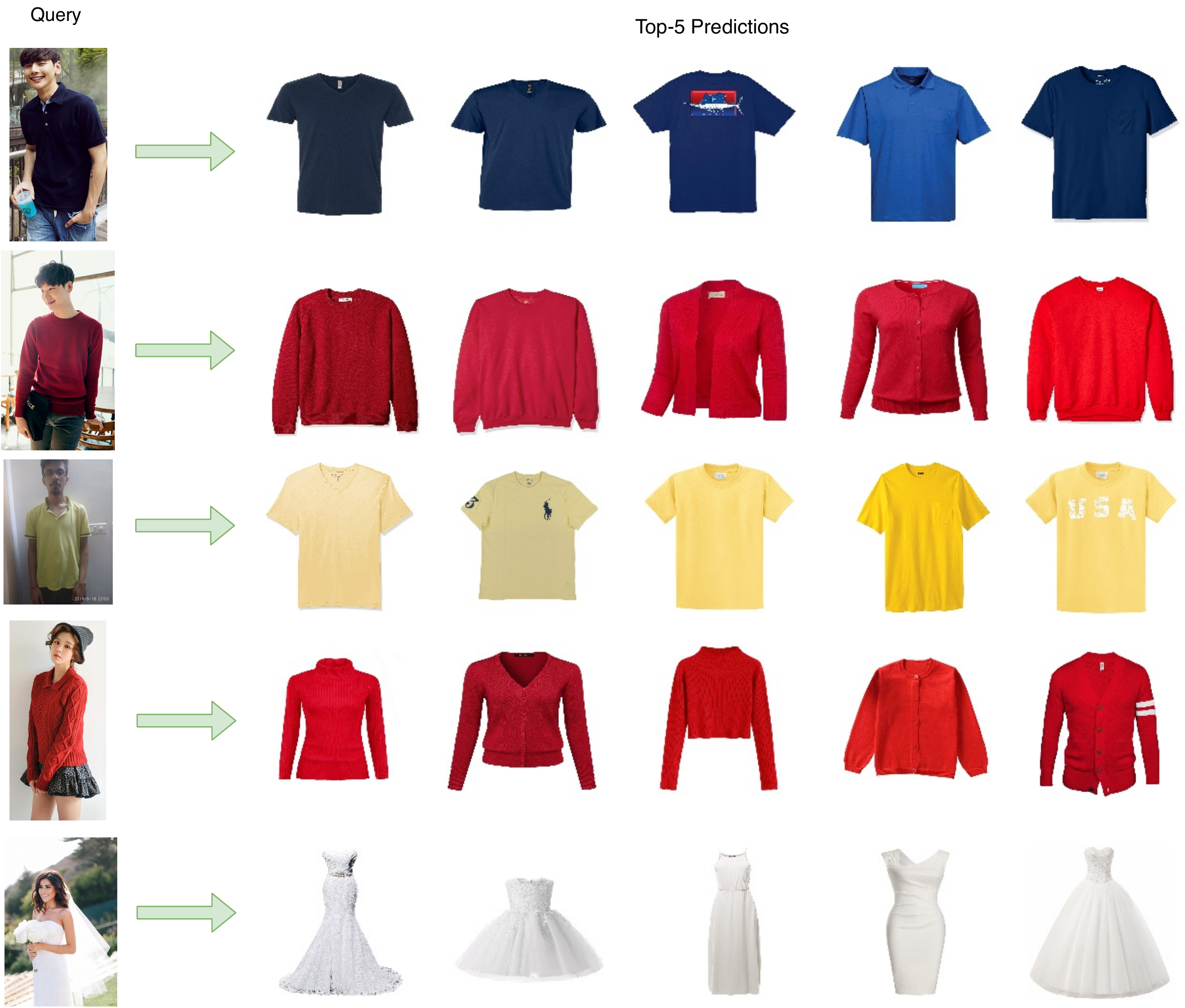}
  \caption{Top 5 predictions of the framework from 5 sample query images.}
  \label{figure:correct_pred}
\end{figure}

Fig~\ref{figure:correct_pred} shows the output produced by our proposed system. It is evident from the performance that the system works well in real-life scenarios as well, where there are significant challenges. Some of these beings pose variance, background clutter, and illumination variance. Moreover, the matching system can match the clothes on multiple attributes simultaneously. It can capture the type of garment as well as other attributes like color and sleeve length. Even the precision values indicate an excellent performance of the matching system, which is visible in Fig.~\ref{figure:correct_pred}.

\begin{figure}[!ht]
\centering
  \includegraphics[width=0.45\linewidth]{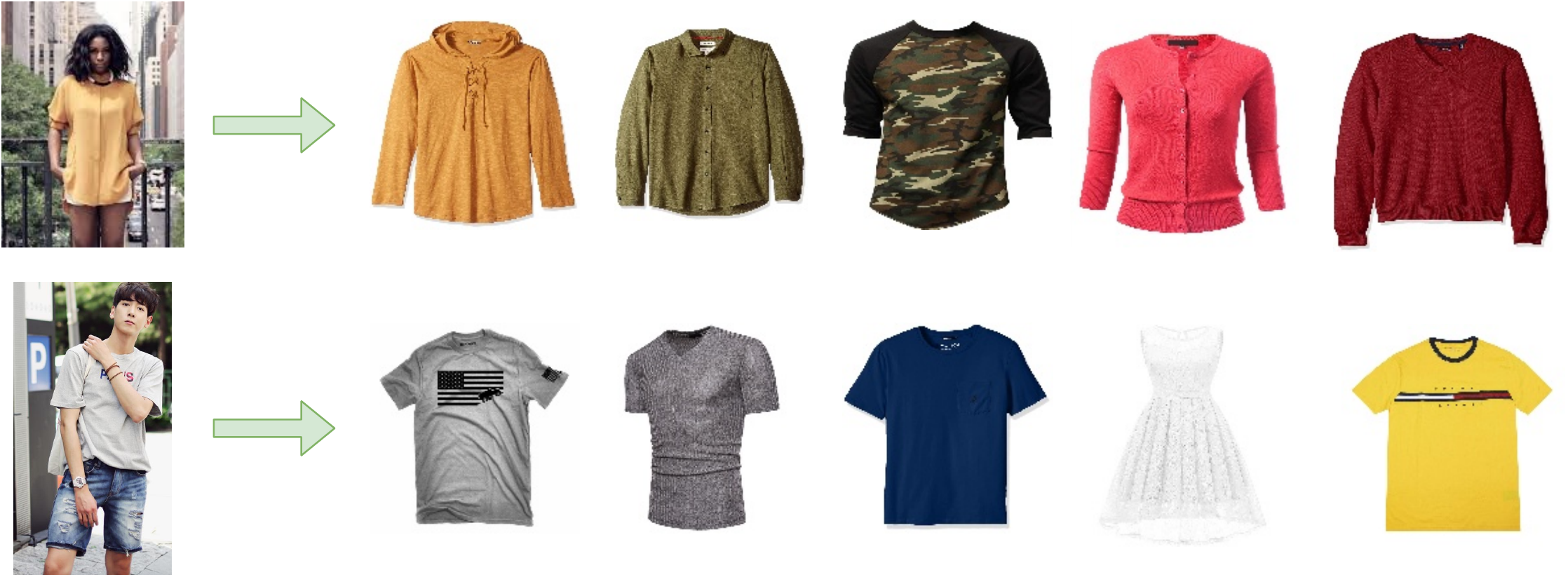}
  \caption{Failure cases for the proposed system.}
  \label{figure:wrong_pred}
  \vspace{-4mm}
\end{figure}

However, the system still suffers some failures. Fig.~\ref{figure:wrong_pred} shows two typical types of failures. The correct matching depends a lot on whether the required garment is in our database or not. If the garment is not in the database, then the system is not able to produce the correct results. The first failure in Fig.~\ref{figure:wrong_pred} shows one such example.

Moreover, since multiple attributes describe a garment, the network sometimes gives importance to some of the attributes than others. In the second failure case, the network can match the type of garment properly that is a half-sleeve t-shirt but is ignoring the color of the shirt. Hence, we need a system in which we could be able to control the priorities of different attributes during search time, also known as attribute matching. 
\section{Conclusion and future work}\label{sec:conc}
In this paper, we presented a new methodology to solve the street to shop problem of finding similar dresses to shop from different photo capturing scenarios. We also introduced a new dataset of clean images of dress shopping products on which we evaluated our proposed methodology. We proposed a methodology in two phases. First, using a GAN to generate clean dress images from daily query photos and then using the generated image as an input to a siamese network in the clothes domain for retrieving similar products. Future work involves expanding the collected dataset to more categories and do an attribute-based matching of products. The matching network should also be able to match generated images with clean product images that contain some parts of the human body part.    

\textbf{Acknowledgements. } We wish to acknowledge and thank Pratyush Gaurav, Shashwat Garg and Sylvia Mittal, students of Indian Institute of Technology, Mandi for their work and contribution to the initial phases of idea and work. 
\bibliographystyle{splncs04}
\bibliography{mybibliography}
\end{document}